\documentclass{article}

\PassOptionsToPackage{numbers, compress}{natbib}

\usepackage[preprint]{nips_2018}

\usepackage[utf8]{inputenc} 
\usepackage[T1]{fontenc}    
\usepackage{hyperref}       
\usepackage{url}            
\usepackage{booktabs}       
\usepackage{amsfonts}       
\usepackage{nicefrac}       
\usepackage{microtype}      
\usepackage{graphicx}
\usepackage{adjustbox}
\usepackage{multirow}
\usepackage{amsmath}

\title{Retraining-Based Iterative Weight Quantization \\for Deep Neural Networks}

\author{
	Dongsoo Lee \\
	Samsung Research \\
    Seoul, Korea \\
  \texttt{dongsoo3.lee@samsung.com} \\
  \And
  Byeongwook Kim \\
  Samsung Research \\
  Seoul, Korea \\
  \texttt{byeonguk.kim@samsung.com} \\
}

\begin{document}

\maketitle

\begin{abstract}
Model compression has gained a lot of attention due to its ability to reduce hardware resource requirements significantly while maintaining accuracy of DNNs.
Model compression is especially useful for memory-intensive recurrent neural networks because smaller memory footprint is crucial not only for reducing storage requirement but also for fast inference operations.
Quantization is known to be an effective model compression method and researchers are interested in minimizing the number of bits to represent parameters.
In this work, we introduce an iterative technique to apply quantization, presenting high compression ratio without any modifications to the training algorithm.
In the proposed technique, weight quantization is followed by retraining the model with full precision weights.
We show that iterative retraining generates new sets of weights which can be quantized with decreasing quantization loss at each iteration.
We also show that quantization is efficiently able to leverage pruning, another effective model compression method.
Implementation issues on combining the two methods are also addressed.
Our experimental results demonstrate that an LSTM model using 1-bit quantized weights is sufficient for PTB dataset without any accuracy degradation while previous methods demand at least 2-4 bits for quantized weights.

\end{abstract}

\section{Introduction}
Since deep neural networks (DNNs) need to support various complex tasks and correspondingly increasing amount of data, the number of parameters of DNNs has increased rapidly.
For example, Deep Speech 2 has 10$\times$ more parameters compared with the previous version \citep{Narang2017}, and the Long Short-Term Memory (LSTM) \citep{LSTM} model using Penn Treebank (PTB) dataset \citep{PTB_google} presents exponentially increasing number of parameters to achieve incremental improvement of test perplexity.
Such a requirement of large DNN model size leads to not only long training time but also high latency and huge storage requirement for performing inference.

Note that the memory accesses dominates the entire inference energy if data reuse is not high enough, since the energy demand of the computation unit is relatively very low \cite{deepcompression}. 
Therefore, model compression is crucial to efficiently implement memory-intensive recurrent neural networks (RNNs) for applications such as natural language processing and speech recognition.
As a result, model compression is actively being studied with the ideas such as low-rank matrix factorization \citep{SVD2013, Xue2013_SVD}, parameter pruning \citep{SHan_2015}, quantization \citep{xu2018alternating}, knowledge distillation \citep{distillation}, and so on. 
Among them, quantization method has numerous advantages of relatively simple implementation, maintaining the model architecture, achieving high compression ratio, and ability to be easily combined with other compression methods.

Quantization techniques widely used for digital signal processing (e.g., uniform quantization \citep{Hubara2016} and balanced quantization \citep{balanced}) are useful for DNNs as well.
In addition, researchers have suggested aggressive binary quantization dedicated to DNNs, where computations are replaced with simple logic units.
BinaryConnect \citep{binaryconnect} learns quantized weights where the forward propagation is performed using quantized weights while full-precision weights are reserved for accumulating gradients.
Scaling factors are stored additionally to compensate for the limited range of binary weights \citep{rastegariECCV16}.
While binary quantization achieves impressive amount of compression, its accuracy in large models (especially for RNNs) is degraded seriously \citep{xu2018alternating}.
Thus, multi-bit quantization techniques based on the underlying principle of reducing mean squared error (MSE) between full-precision and quantized weights are introduced to maintain the accuracy.
For example, Alternating multi-bit quantization \citep{xu2018alternating} extends the XNOR-Net architecture \citep{rastegariECCV16} to find a set of coefficients minimizing MSE, and demonstrates that 2-4 bits for parameters are enough to meet the accuracy of RNNs.
K-means clustering reducing MSE and the loss has also been shown to maintain the original model accuracy \citep{limitquant2017}.
Even though quantization can be performed as an additional step after training is finished \citep{deepcompression, adaptive}, most of recent studies modify training procedures to incorporate quantization effect to mitigate the accuracy degradation at the cost of additional training algorithm design considerations \citep{limitquant2017, binaryconnect, rastegariECCV16, wu2018, xu2018alternating, Dorefa, ternary2017}.

In this work, we introduce two novel quantization techniques reducing the number of quantization bits significantly without modifying underlying training procedures.
The first technique is \textit{iterative quantization method} where quantization is conducted repeatedly after retraining the model with full precision.
We find new local minima in DNN search space which provide gradually decreasing amount of quantization loss.
As a result, accuracy is improved with more retraining iterations given the same number of quantization bits.
The second technique is efficiently \textit{merging pruning and quantization}. 
Our experimental results report the improvement in quantization after the number of parameters to be quantized is reduced by pruning technique. 
In consequence, the weights of LSTM model on the PTB dataset can be quantized to only 1-bit with the accuracy of full-precision network.

\section{Related Work}
In this section, we summarize previous quantization methods which we utilize as a baseline.

Following the Binary-Weight-Networks \citep{rastegariECCV16}, a weight vector $\mathbf{w}$ is approximated to be $\alpha \mathbf{b}$ by using a scaling factor $\alpha$ and a vector $\mathbf{b}\left(= \left\{-1,+1\right\}^n\right)$, where $n$ is the vector size.
Then $\Arrowvert \mathbf{w}-\alpha\mathbf{b}\Arrowvert^2$ is minimized to obtain 
\begin{equation} \label{eq:1}
\mathbf{b}^* = {\rm sign}(\mathbf{w}),\; \alpha^* = \frac{\mathbf{w}^{\rm T} \mathbf{b}^*}{n}.
\end{equation}

1-bit quantization shown in Eq. (\ref{eq:1}) is extended to multi-bit ($k$-bit) quantization using a greedy method \citep{Greedy_Quan} where the $i^{\rm th}$-bit ($i > 1$) quantization is performed by minimizing the residue of $(i-1)^{\rm th}$-bit quantization as following:
\begin{equation} \label{eq:2}
\min_{\alpha_i ,\mathbf{b}_i} \Arrowvert \mathbf{r}_{i-1} - \alpha_i \mathbf{b}_i \Arrowvert ^2, \; {\rm where} \; \mathbf{r}_{i-1} = \mathbf{w} - \sum_{j=1}^{i-1} \alpha_j \mathbf{b}_j, \; 1 < i \le k.
\end{equation}
The optimal solution of Eq. (\ref{eq:2}) is given as
\begin{equation} \label{eq:3}
\mathbf{b}_i^* = {\rm sign}(\mathbf{r}_{i-1}), \; \alpha_i^* = \frac{\mathbf{r}_{i-1}^{\rm T} \mathbf{b}_i^*}{n}.
\end{equation}
Note that Eq. (\ref{eq:3}) is not the optimal solution for $\Arrowvert \mathbf{w} - \sum_{i=1}^k \alpha_i \textbf{b}_i \Arrowvert$.
As an attempt to lower quantization error, $\left\{ \alpha_i\right\}_{i=1}^k$ can be refined as $\left[ \alpha_1, ..., \alpha_k \right] = \left( \left( \mathbf{B}_k^{\rm T} \mathbf{B}_k \right)^{-1} \mathbf{B}_k^{\rm T} \mathbf{w} \right)^{\rm T}$, when $\mathbf{B}_k = \left[ \mathbf{b}_1, ... \mathbf{b}_k \right]$  \citep{Greedy_Quan}. 
Further improvement can be obtained by using Alternating multi-bit method \citep{xu2018alternating}, where $\mathbf{B}_k$ is obtained by binary search given a new refined $\left\{ \alpha_i\right\}_{i=1}^k$, and $\left\{ \alpha_i\right\}_{i=1}^k$ and $\mathbf{B}_k$ are refined alternatively.
This procedure is repeated until there is no noticeable improvement in MSE.
Quantized weights are used and updated during training procedures associated with special considerations on quantized weights such as weight clipping and ``straight-through estimate'' \citep{xu2018alternating}.

Quantization using binary weights simplifies the matrix operations by significantly reducing the number of multiplications \citep{xu2018alternating}.
It also eliminates the need for dequantization for inference, leading to reduced on-chip memory size for weights.
K-means clustering is another effective quantization method even though the algorithmic complexity is higher than binary codes quantization in general \citep{limitquant2017}.
Some researchers have also suggested Hessian-weighted measures over MSE in order to represent accuracy at the cost of additional computational complexity.
In this work, Alternating quantization \citep{xu2018alternating} is used as a baseline even though our proposed method can be combined with other quantization techniques as well.

\section{Iterative Quantization Using Full-Precision Retraining}

Quantization loss and accuracy of DNNs after weight quantization depend on the smoothness on the loss surface after training \citep{flat_minima}.
Flat minima reduce accuracy loss induced by quantization since DNNs become less sensitive to the variation of weights \citep{binaryconnect,flat_minima}.
In order to reduce the number of quantization bits, it is important to perform full-precision retraining to reach flatter minima where the accuracy gap between full-precision weights and quantized weights is diminished.
In this section, we present our proposed iterative quantization scheme to reduce the number of bits required for quantization and discuss experimental results.

\subsection{Motivation and the Proposed Method}

\begin{figure}[t]
	\begin{minipage}{.52\textwidth}
		\includegraphics[width=1\linewidth]{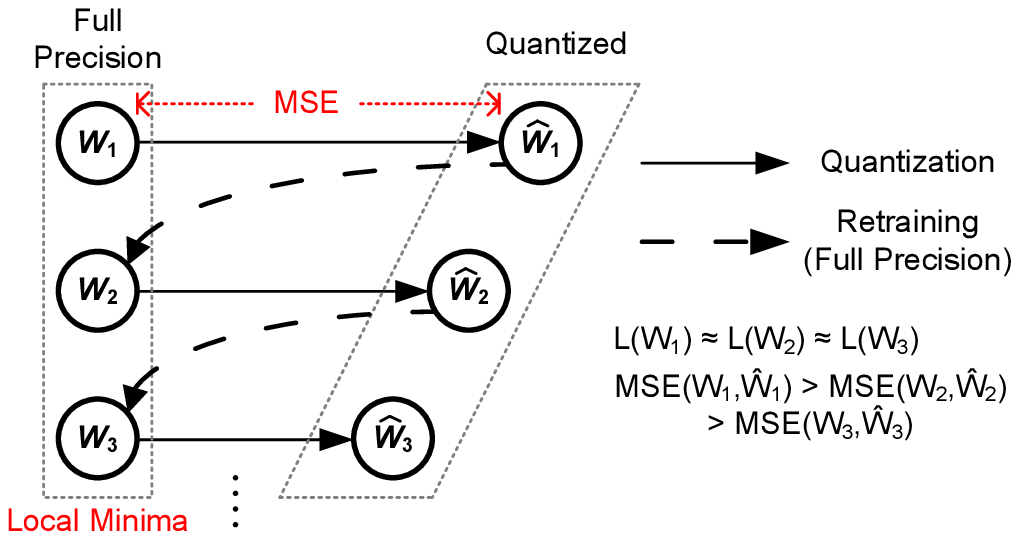}
		\caption{The proposed iterative quantization scheme with full-precision retraining while searching for better local minima for quantization.}
		\label{fig:motivation}
	\end{minipage}
\hfill
	\begin{minipage}{.43\textwidth}
		\includegraphics[width=1\linewidth]{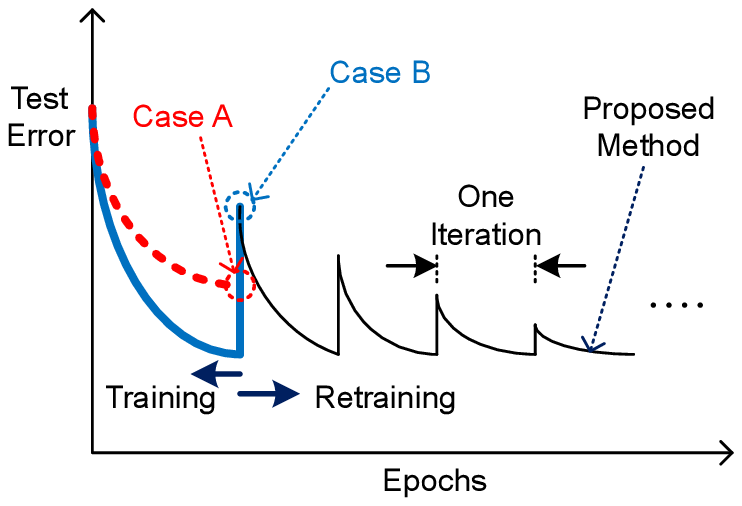}
		\caption{Quantization scheme comparison where Case A presents quantization-aware training method and Case B describes a one-time quantization after training.}
		\label{fig:quant_comparison}
	\end{minipage}
\end{figure}

Most of the local minima have similar loss function values to the global optimum for high-dimensional DNNs \citep{nature}.
Nonetheless, local minima with similar DNN accuracy can represent vastly different quantization loss depending on the smoothness of the loss surface \citep{loss_surface}.
Correspondingly, searching for the local minima employing the minimum quantization loss is the key to minimizing the number of quantization bits.
Such an effort can be expressed by the following equation
\begin{equation} \label{eq:4}
\operatorname*{argmin}_\mathbf{w} \rm{MSE}\left( \mathbf{w}, \mathbf{\hat{w}} \right) = \operatorname*{argmin}_\mathbf{w} \Arrowvert \mathbf{w} - \mathbf{\hat{w}} \Arrowvert ^2 \quad \rm{s.t.} \quad \it{L}(\mathbf{w}) < \varepsilon ,
\end{equation}
where $\mathbf{w}$ is quantized to $\mathbf{\hat{w}}$, $L(\mathbf{w})$ is the loss function of weight set $\mathbf{w}$, and $\varepsilon$ is the maximum loss value for the local minima.

The first step of our proposed technique is to quantize $\mathbf{w_1}$, which is the weight set after full-precision training, to obtain the quantized weight set $\hat{\mathbf{w_1}}$.
$\hat{\mathbf{w_1}}$ is then dequantized and used as the initial weights for retraining to obtain $\mathbf{w_2}$ representing a different local minimum.
Note that new local minima to find $\mathbf{w_2}$ are explored in the area close to the point corresponding to the previously quantized weights $\hat{\mathbf{w_1}}$.
If the difference between $\mathbf{\hat{w_1}}$ and $\mathbf{\hat{w_2}}$ is not large while $\mathbf{w_2}$ is much closer to $\mathbf{\hat{w_1}}$ than $\mathbf{w_1}$, then ${\rm MSE}\left(\mathbf{w_2, \hat{w_2}} \right)$ is smaller than ${\rm MSE}\left(\mathbf{w_1, \hat{w_1}} \right)$.
Iterating such a procedure (i.e., $\mathbf{\hat{w_n}}$ is dequantized and retraining is performed to obtain $\mathbf{w_{n+1}}$ at the $\mathbf{n}^{\rm{th}}$ iteration) continuously reduces MSE of full-precision $\mathbf{w_n}$ and quantized $\mathbf{\hat{w_n}}$.
Moreover, adding noise to weight through this quantization procedure enables the model to reach flatter minima \citep{distillation2018} along with retraining which induces a robustness to the weight variation.
Consequently, because of reduced $\rm{MSE}$ and flatter minima, such iterations gradually improve the DNN accuracy using quantized weights and reduce the number of quantization bits (see Figure 9 in Appendix).
Throughout the iterations, different local minima with full-precision exhibit similar loss values $\left( L(\mathbf{w_1}) \simeq L(\mathbf{w_2}) \simeq L(\mathbf{w_3}) \right)$, while a reduction in loss of quantized weights is observed as $L(\mathbf{\hat{w_1}}) > L(\mathbf{\hat{w_2}}) > L(\mathbf{\hat{w_3}}) $ as shown in Figure \ref{fig:motivation}.

Case B in the Figure \ref{fig:quant_comparison} represents one-time quantization conducted as a post-processing after full-precision training (e.g., \citep{Greedy_Quan,deepcompression}).
Case A in the Figure \ref{fig:quant_comparison}, on the other hand, requires a modified training algorithm aiming at improved quantization quality by taking into account quantization process for the forward and backward propagation (e.g., \citep{limitquant2017, binaryconnect, rastegariECCV16, wu2018, xu2018alternating, Dorefa, ternary2017}).
Despite thoroughly designed quantization-dedicated training algorithms for Case A, the amount of weight update during training is not large enough to escape from a certain local minimum \citep{resnet_quan}.
Hence, previous quantization methods lack the ability to fulfill Eq. \ref{eq:4} which should be supported by exploring various local minima in the search space.
The proposed iterative quantization inherits the advantage of Case B (no modification to training procedure to incorporate quantization) and achieves smaller quantization loss than Case A.

\subsection{Experimental Results}

\begin{table}
	\caption{Test perplexity on PTB dataset with various number of quantization bits for LSTM weights without retraining iterations.}
	\label{table:PTB_Alternating}
	\centering
	\begin{tabular}{c|c c c c c c c}
		\hline
		Model Size & Full & 1-bit & 2-bit & 3-bit & 4-bit & 5-bit & 6-bit \\
		\hline
		Small & 115.111 & 426.796 & 126.427 & 116.658 & {\bf 114.878} & 114.677 & 114.743 \\
		Medium & 83.571 & 240.388 & 86.590 & {\bf 83.139} & 82.806 & 82.939 & 83.028 \\
		Large & 78.275 & 163.381 & 79.424 & {\bf 76.945} & 77.045 & 77.424 & 77.592 \\
		\hline
	\end{tabular}
\end{table}

\begin{figure}
	\begin{minipage}{.495\textwidth}
		\includegraphics[width=1\linewidth]{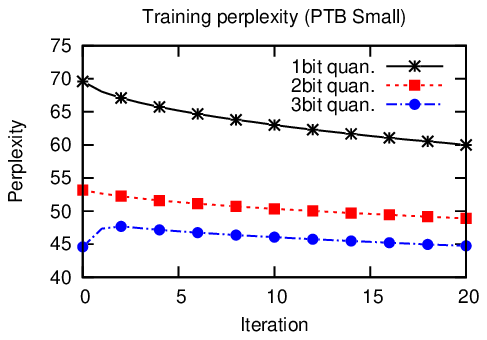}
	\end{minipage}
	\hfill
	\begin{minipage}{.495\textwidth}
		\includegraphics[width=1\linewidth]{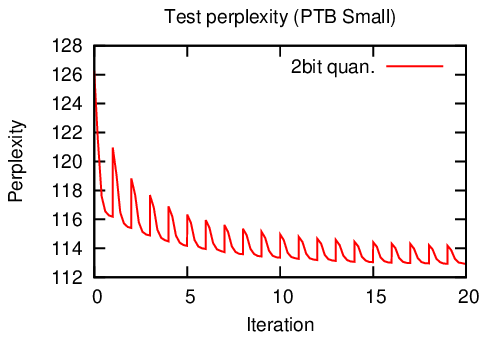}
	\end{minipage}
	\newline
	\begin{minipage}{.495\textwidth}
		\includegraphics[width=1\linewidth]{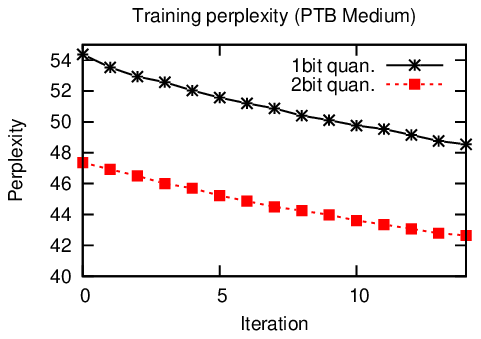}
	\end{minipage}
	\hfill
	\begin{minipage}{.495\textwidth}
		\includegraphics[width=1\linewidth]{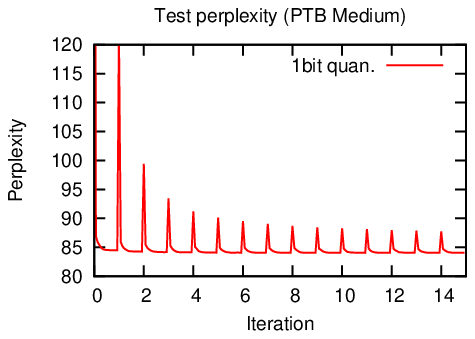}
	\end{minipage}
	
	\caption{Training and test perplexity with full-precision retraining at each iteration using our proposed iterative quantization techniques. See Appendix for more graphs.}
    \label{fig:perplexity_nopruning}
\end{figure}

There exist many attempts to compress convolutional neural networks (CNNs) while RNNs have received less attention \citep{xu2018alternating}.
We chose LSTM model \citep{PTB_google} with PTB dataset \citep{Marcus_1993} to verify our proposed techniques even though our proposed technique is a general one.
Following the model given in \citep{PTB_google}, there are 2 LSTM layers and 3 configurations depending on the number of LSTM units in a layer (small, medium, and large models have 200, 650, and 1500 LSTM units, respectively).
We follow the learning schedule in \citep{PTB_google} for retraining except that the initial learning rate is reduced by 100.
The number of epochs for retraining is the same as that of the training.
The accuracy is measured by Perplexity Per Word (PPW), referred as simply perplexity in this paper.
Table \ref{table:PTB_Alternating} presents the accuracy of 3 configurations after applying Alternating multi-bit quantization \citep{xu2018alternating} \footnote{In this work, `Alternating' quantization does not contain activation quantization and training algorithm modification unless stated otherwise.} where quantization is independently conducted for each row \footnote{The quantization table values $\alpha$ are in single-precision floating-point format while binary codes of each weight have 1-6 bits.}.
In \citep{xu2018alternating}, even though the training procedure is carefully re-designed and activations are not quantized, at least 4 bits are required for weight quantization with 300 LSTM units to achieve full precision accuracy.
As Alternating quantization is known to be efficient for RNNs, we utilize it as a baseline in this work to investigate how many quantization bits can be reduced further (compared with Table \ref{table:PTB_Alternating}) by using our proposed iterative method.

Figure \ref{fig:perplexity_nopruning} shows training perplexity (left) after each full-precision retraining and test perplexity (right) sampled during retraining and each quantization round.
For both small and medium PTB models, training successfully converges for iterations and training perplexity decreases as we utilize more bits for quantization.
The test perplexity transitions from our experiments (Figure \ref{fig:perplexity_nopruning}) matches with the earlier expectations (Figure \ref{fig:quant_comparison}).
Note that the test perplexity at iteration '0' is equivalent to the result of Alternating quantization without iterations.
It is obvious that iterating full-precision retraining and quantization significantly reduce test perplexity for a given number of quantization bits.
In Figure \ref{fig:perplexity_nopruning} for test perplexity on the small PTB model, quantization is conducted before retraining accomplishes full precision accuracy.
Similarly to gradual pruning \citep{Narang2017}, test perplexity is continuously improved through iterations even when retraining is finished early for each iteration.

\begin{figure}
	\begin{minipage}{.495\textwidth}
		\includegraphics[width=1\linewidth]{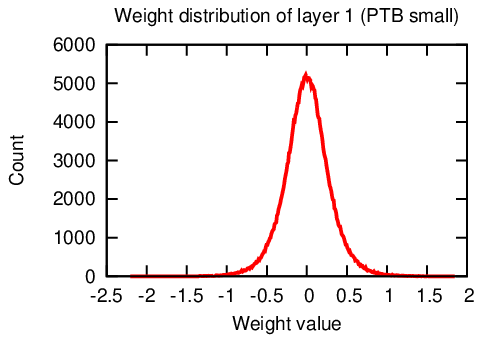}
		
	\end{minipage}
	\hfill
	\begin{minipage}{.495\textwidth}
		\includegraphics[width=1\linewidth]{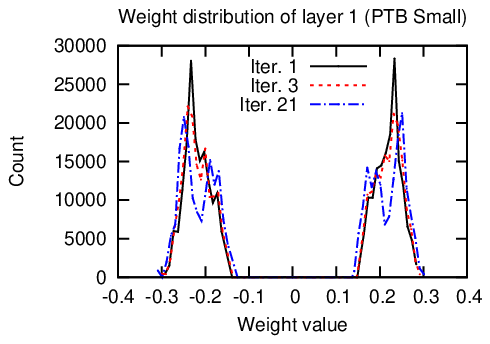}
	\end{minipage}
	\caption{Weight distribution of layer 1 after training (left) and after iterative quantization (right). Long tails on the left side disappear on the right side due to quantization. See Appendix for layer 2.}
	\label{fig:weight_dist_nopruning_layer1}
\end{figure}

\begin{figure}
	\begin{minipage}{.495\textwidth}
		\includegraphics[width=1\linewidth]{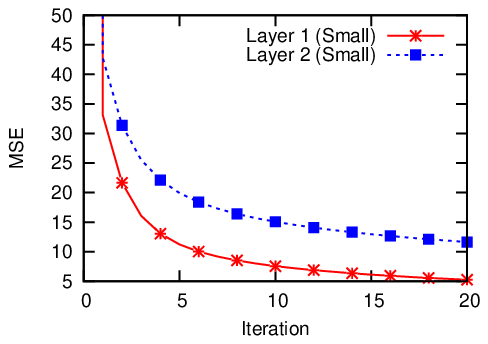}
		
	\end{minipage}
	\hfill
	\begin{minipage}{.495\textwidth}
		\includegraphics[width=1\linewidth]{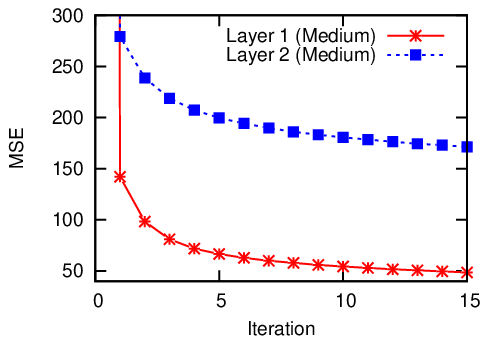}
	\end{minipage}
	\caption{${\rm MSE}\left(\mathbf{w_n, \hat{w_n}} \right)$ at each iteration using 1-bit quantization on the PTB dataset using LSTM small (left) and medium (right) models.}
	\label{fig:MSE_nopruning}
\end{figure}

Figure \ref{fig:weight_dist_nopruning_layer1} presents weight distributions of layer 1 right after training procedure (left) and at the 1st, 3rd, and 21st 1-bit quantization round after corresponding retraining (right).
We can observe that the changes in quantized weight distributions become smaller during iterations.
This means $\mathbf{\hat{w_n}}$ converges to a certain value ($\mathbf{\hat{w_\infty}}$) as $\mathbf{n}$ increases.
Based on the discussions in Section 3.1, if we manipulate full-precision weights at local minima ($\mathbf{\hat{w_n}}$) through iterations to be closer to $\mathbf{\hat{w_\infty}}$, then ${\rm MSE}\left(\mathbf{w_n, \hat{w_n}} \right)$ deceases as $n$ increases.

Indeed, ${\rm MSE}\left(\mathbf{w_n, \hat{w_n}} \right)$ is reduced at every iteration as shown in Figure \ref{fig:MSE_nopruning}.
In the case of Alternating 1-bit quantization on the PTB small model, ${\rm MSE} \left(={\rm MSE}\left(\mathbf{w_1, \hat{w_1}} \right)\right)$ values are 11181.5 and 17972.6 for layer 1 and layer 2, respectively (on the PTB medium model, $\rm{MSE}$ values are 13683.1 and 27552.1 for layer 1 and layer 2, respectively).
The proposed method, hence, dramatically reduces ${\rm MSE}$ and test perplexity for the same number of quantization bits compared with Alternating quantization method.

\begin{table}
	\caption{Test perplexity on the PTB small model with various number of quantization tables per row (without retraining). We assume that $\alpha$ values in Eq. \ref{eq:2} are represented by using 16 bits. The size of quantized weights $S$ does not include the table size.}
	\label{table:multitable}
	\centering
	\begin{tabular}{c | c c | c c | c c}
		\hline
		Number of & \multicolumn{2}{|c|}{\underline{1-bit ($S$=39.07KB)}} & \multicolumn{2}{|c|}{\underline{2-bit ($S$=78.13KB)}} & \multicolumn{2}{|c}{\underline{3-bit ($S$=117.19KB)}} \\
		Tables & Test & Table & Test & Table & Test & Table \\
		Per Row & Perplexity & Size & Perplexity & Size & Perplexity & Size \\
		\hline
		1 & 426.796 & 0.78KB & 126.427 & 1.56KB & 116.658 & 2.34KB \\
		2 & 317.928 & 1.56KB & 124.936 & 3.13KB & 116.086 & 4.69KB \\
		4 & 240.856 & 3.13KB & 121.623 & 6.25KB & 115.651 & 9.38KB\\
		8 & 238.891 & 6.25KB & 120.958 & 12.5KB & 115.232 & 18.8KB \\
		\hline
	\end{tabular}
\end{table}

Besides iterations, we conducted experiments to investigate whether more number of independent quantization tables per row results in better quantization quality.
As shown in Table \ref{table:multitable}, employing more quantization tables obviously improves test perplexity as ${\rm MSE}$ of Eq. \ref{eq:2} becomes smaller due to the shrinking number of weights to be quantized for each quantization table.
Such an improvement on test perplexity, however, requires exponentially increasing quantization table size (=number of total $\alpha$ values $\times 16$ if 16 bits are used for $\alpha$).
Therefore, for the next sections, only one quantization table per row is used.

\section{Combining Quantization and Pruning}

In this section, we empirically study the impact of pruning weights on quantization.
Intuitively, quantization should be able to leverage high pruning rate.
First, because MSEs of pruned weights are 0, total ${\rm MSE}\left(\mathbf{w_n, \hat{w_n}} \right)$ decreases and the test perplexity using quantized weights is improved correspondingly.
In addition, pruning can be recognized as a regularization technique \citep{DSD} which facilitates flatter minima.
Most of prior efforts of combining pruning and quantization are given in the form of ternary weight quantization \citep{ternary2017}.
For example, if magnitude of a weight is smaller than a certain threshold value, the corresponding weight is replaced with `0' in either deterministic or stochastic way, while the other weights can be either `-1' or `+1'.

\begin{figure}
	\begin{minipage}{.495\textwidth}
		\includegraphics[width=1\linewidth]{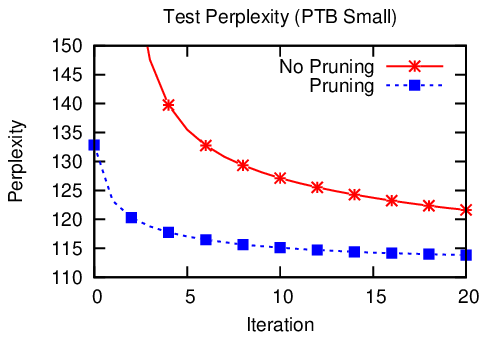}
		
	\end{minipage}
	\hfill
	\begin{minipage}{.495\textwidth}
		\includegraphics[width=1\linewidth]{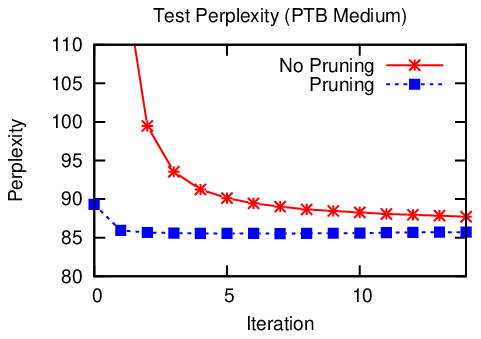}
	\end{minipage}
	\caption{1-bit quantization test perplexity with pruning 80\% of weights and without pruning on the PTB small (left) and medium (right) models. Pruning weights reduces retraining iterations significantly for the same target test perplexity.}
	\label{fig:pruning_effect_comparison}
\end{figure}

When the weights are allowed to have one of three possible choices, then the combination of pruning and 1-bit quantization requires minimum 2 bits for weights.
In order to achieve less than 1 bit on average after quantization and pruning, minimizing the amount of masking information followed by pruning is crucial and should be much less than the size of binary masking used in ternary quantization.
Widely used sparse matrix formats, such as Compressed Sparse Row (CSR), are not efficient for low-bit quantization because of relatively huge index information for each non-zero weight.
We can efficiently compress index information after pruning using recently studied Viterbi-based index compression technique \citep{lee2018viterbibased}.
For Viterbi decompressor \citep{lee2018viterbibased}, we chose $\it{NUM_v}$=50, $\it{NUM_c}$=5, and $R$=10 without skip state for 80\% pruning rate on the PTB small and medium models.
Such a configuration compresses binary masking information by 90\%, resulting in 0.1 bits/weight for index.
Since 0.2 bits/weight is required to represent non-zero weights (storage requirement for quantization table $\alpha$ values is negligible), we need 0.3 bits/weight in total, which is 6.7$\times$ compression compared with 2-bit ternary quantization.
Hence, an efficient sparse matrix format, such as VCM format \citep{lee2018viterbibased}, is the key for high compression rate after combining pruning and quantization.

Figure \ref{fig:pruning_effect_comparison} describes test perplexity when 1-bit quantization is conducted after pruning 80\% of weights (following the context of magnitude-based pruning \citep{SHan_2015}).
As shown in Figure \ref{fig:pruning_effect_comparison}, the leverage effect of pruning and 1-bit quantization rapidly improves test perplexity to be saturated to the full-precision results (115.111 for small model and 83.571 for medium model).
In other words, pruning reduces the number of iterations for a certain target accuracy, resulting in fast retraining of the proposed technique.
Note that the accuracy obtained by pruning and iterative quantization even exceeds that of full-precision training due to the regularization effect of pruning and quantization \citep{DSD,xu2018alternating}.
Combination of pruning and 1-bit quantization for the proposed method is not a simple coupling but a way to maximize the synergy effect.

\begin{figure}
	\begin{minipage}{.495\textwidth}
		\includegraphics[width=1\linewidth]{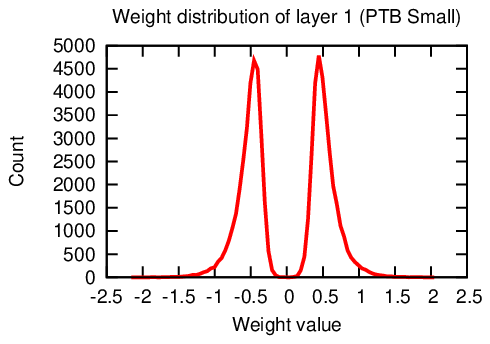}
		
	\end{minipage}
	\hfill
	\begin{minipage}{.495\textwidth}
		\includegraphics[width=1\linewidth]{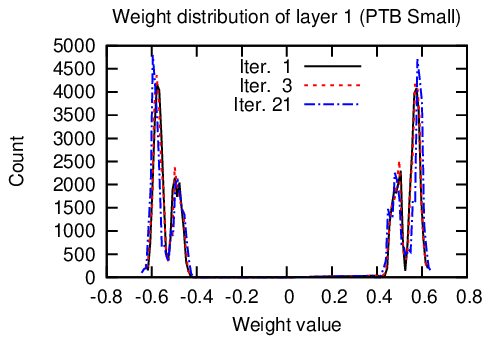}
	\end{minipage}
	\caption{Weight distribution of layer 1 on the PTB small model after pruning 80\% of weights and retraining (left) and after additional 1-bit iterative quantizaton (right). See Appendix for layer 2.}
	\label{fig:weight_dist_pruning_layer1}
\end{figure}

\begin{figure}
	\begin{minipage}{.495\textwidth}
		\includegraphics[width=1\linewidth]{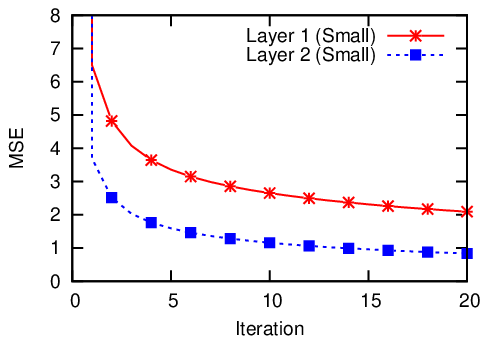}
		
	\end{minipage}
	\hfill
	\begin{minipage}{.495\textwidth}
		\includegraphics[width=1\linewidth]{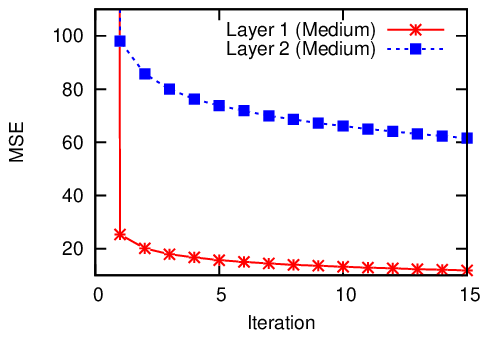}
	\end{minipage}
	\caption{${\rm MSE}\left(\mathbf{w_n, \hat{w_n}} \right)$ at each iteration using 80\% pruning scheme and additional 1-bit iterative quantization on the PTB small (left) and medium (right) models.}
	\label{fig:MSE_pruning}
\end{figure}

Compared with Figure \ref{fig:weight_dist_nopruning_layer1}, quantized weights after pruning yield less amount of change in weights at different iterations as described in Figure \ref{fig:weight_dist_pruning_layer1} .
In other words, as a consequence of pruning, $\mathbf{\hat{w_n}}$ converges to $\mathbf{\hat{w_\infty}}$ faster as $\mathbf{n}$ increases.
We conjecture that iterative quantization using less number of weights (due to pruning) finds flatter minima more quickly, and thus $\mathbf{\hat{w_n}}$ achieves faster convergence speed.
${\rm MSE}\left(\mathbf{w_n, \hat{w_n}} \right)$ through iterations after pruning (in Figure \ref{fig:MSE_pruning}) also decreases more rapidly with much smaller final values compared with the case of preserving all the weights (Figure \ref{fig:MSE_nopruning}).
In sum, the advantage of the proposed iterative quantization is boosted by pruning weights before performing quantization process.

Table \ref{table:all_results} summarizes the results discussed so far on the PTB small model.
While Alternating quantization method demands at least 4 bits to meet or exceed the full-precision test perplexity (115.111), the proposed method (pruning + iterative quantization) requires only 1-bit for quantization.
For the other PTB models, pruning and iterative quantization achieve the test perplexity of 85.540 (on the PTB medium model with 80\% pruning rate) and 80.308 (on the PTB large model with 85\% pruning rate) which are close to the full-precision results with 1-bit quantization.

\begin{table}
	\caption{Test perplexity comparison on the PTB small model. Text perplexity after full-precision training is \bf{115.111}}
	\label{table:all_results}
	\centering
	\begin{tabular} {l|c c c c}
		\hline
		Method & 1-bit & 2-bit & 3-bit & Comment \\
		\hline
		Alternating \citep{xu2018alternating} & 426.796 & 126.427 & 116.658 & Full-precision Training \\
		Alternating + Multi-table & 238.891 & 120.958 & 115.232 & 8 Tables/row\\
		Iterations(Alt.+Retraining) & 119.162 & 114.137 & 114.713 & 21 Iterations\\
    	Pruning+Iter.(Alt.+Retraining) & 113.426 & 111.113 & --- & 80\% Pruning + 21 Iter.\\
		\hline
	\end{tabular}
\end{table}

\begin{table}
	\caption{Test perplexity comparison on Text8 dataset. Test perplexity after full-precision training is \textbf{105.965}.}
    \label{table:text8}
    \centering
    \begin{tabular} {l|c c c c c c}
    	\hline
    	Method & 1-bit & 2-bit & 3-bit & 4-bit & 5-bit & 6-bit \\
    	\hline
    	Alternating & 370.808 & 131.888 & 114.174 & 110.993 & 110.211 & 109.972 \\
        Pruning+Iter.(Alt.+Retraining) & 123.666 & 109.196 & 106.824 & 106.126 & 105.828 & 105.632 \\
    	\hline
    	\end{tabular}
\end{table}

We tested our proposed technique using larger dataset Text8 corpus\citep{Text8} and the LSTM model introduced in \citep{Xie_2017}.
We follow the same training, validation, and test setting, data preprocessing, and hyperparameters as in \citep{Xie_2017} except the batch size (100) and the hidden layer size (1024) which are introduced as per \citep{xu2018alternating} for the ease of accuracy comparison.
Table \ref{table:text8} presents test perplexity of our proposed quantization (3 retraining iterations with 90\% pruning rate) and Alternating quantization.
Our proposed quantization scheme achieves much improved DNN accuracy for the same model setting and the number of quantization bits.
Note that the model \citep{Xie_2017} includes additional hyperparameters for new regularization techniques.
It would be necessary to tune those additional hyperparameters for our proposed quantization scheme in order to achieve even lower test perplexity than those in Table \ref{table:text8}.

In the proposed quantization framework, improved quantization quality is mainly based on finding flatter minima.
Hence, besides quantization, it would be interesting to investigate the characteristics of full-precision weights after iterations and study their impact on generalization capability.

\section{Conclusion}
In this work, we have introduced an iterative quantization technique based on full-precision retraining that continuously diminishes the difference between full precision weights and quantized weights (MSE) significantly.
Retraining finds flatter minima through iterative quantization (recognized as noise insertion process), and hence quantized weights obtain less loss function values even for the same MSE.
Due to reduced MSE and flatter minima through iterations, the proposed iterative quantization method reduces the number of quantization bits.
We have also demonstrated that pruning weights can be combined with the proposed iterative quantization scheme, dramatically reducing the amount of retraining and the required number of bits for quantization further.
On the PTB LSTM models, we achieved the full-precision test perplexity using only 1-bit weight quantization.

\small
\bibliography{./dongsoo_2018_nips_quantization}

\newpage
\appendix
\section{Appendix}


\begin{figure}[h]
\centering
	\includegraphics[width=0.6\linewidth]{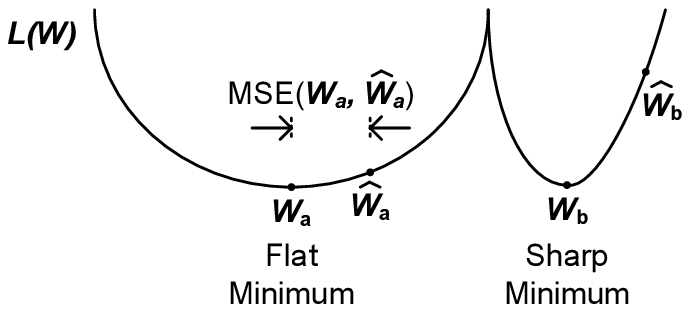}
    \caption{The loss function $L(\mathbf{\hat{w_n}})$ has a lower value when ${\rm MSE}\left(\mathbf{w_n, \hat{w_n}} \right)$ decreases after quantization. Even if ${\rm MSE}\left(\mathbf{w_a, \hat{w_a}} \right)$ = ${\rm MSE}\left(\mathbf{w_b, \hat{w_b}} \right)$ in this figure, we obtain $L(\mathbf{\hat{w_a}}) < L(\mathbf{\hat{w_b}})$ when $\mathbf{W_a}$ exists at a flat minimum. Hence, both the loss surface and ${\rm MSE}$ significantly affect quantization quality.}
    \label{fig:flat_minima_appen}
\end{figure}

\begin{figure}[h]
	\begin{minipage}{.495\textwidth}
		\includegraphics[width=1\linewidth]{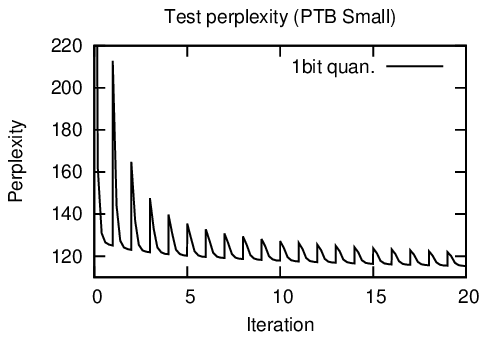}
	\end{minipage}
	\hfill
	\begin{minipage}{.495\textwidth}
		\includegraphics[width=1\linewidth]{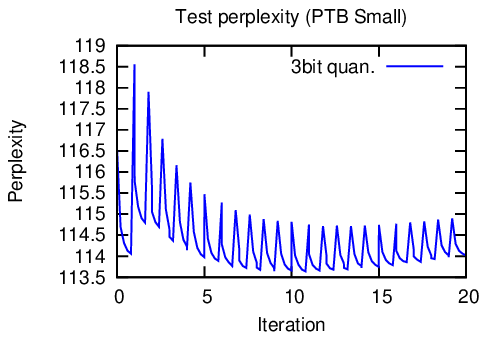}
	\end{minipage}
	\caption{Test perplexity on the PTB small model throughout iterations using our proposed iterative quantization scheme using 1-bit (left) and 3-bit (right) weight quantization.}
	\label{fig:test1}
\end{figure}

\begin{figure}[h]
	\centering
	\includegraphics[width=0.495\linewidth]{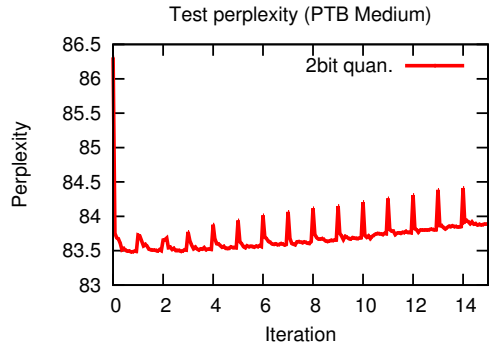}
	\caption{Test perplexity on the PTB medium model throughout iterations using our proposed iterative quantization scheme using 2-bit weight quantization. Continuing iterative quantization even after reaching the accuracy of full-precision training hurts the test perplexity due to overfitting.}
	\label{fig:appendix_medium_test}
\end{figure}

\begin{figure}[h]
	\begin{minipage}{.495\textwidth}
		\includegraphics[width=1\linewidth]{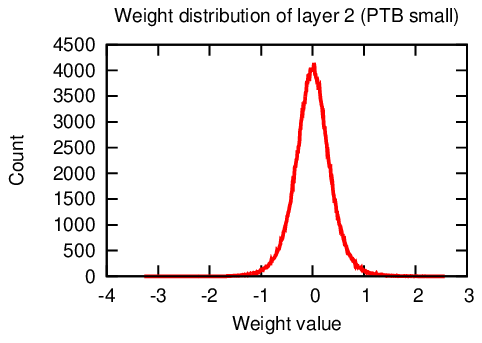}
		
	\end{minipage}
	\hfill
	\begin{minipage}{.495\textwidth}
		\includegraphics[width=1\linewidth]{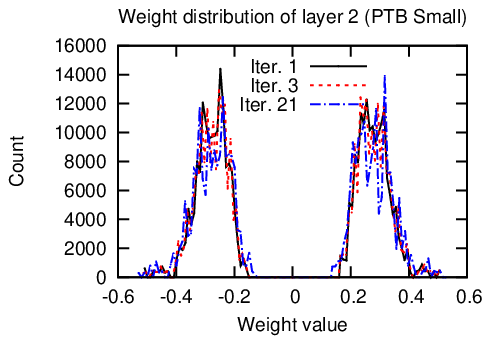}
	\end{minipage}
	\caption{Weight distribution of layer 2 on the PTB small model after training (left) and after iterative quantizatons (right) without pruning weights.}
	\label{fig:weight_dist_nopruning_layer2}
\end{figure}

\begin{figure}[h]
	\begin{minipage}{.495\textwidth}
		\includegraphics[width=1\linewidth]{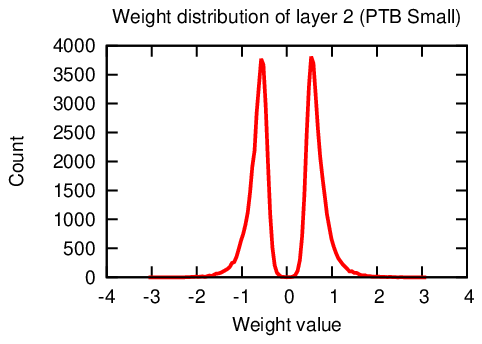}
		
	\end{minipage}
	\hfill
	\begin{minipage}{.495\textwidth}
		\includegraphics[width=1\linewidth]{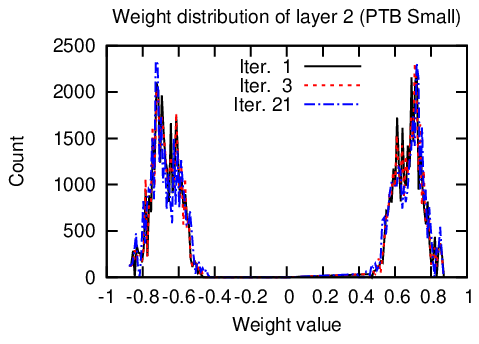}
	\end{minipage}
	\caption{Weight distribution of layer 2 on the PTB small model after pruning 80\% weights and retraining (left) and after additional iterative quantizaton (right).}
	\label{fig:weight_dist_pruning_layer2}
\end{figure}

\end{document}